\pdfoutput=1

\documentclass[11pt]{article}

\usepackage[]{ACL2023}

\usepackage{times}
\usepackage{latexsym}

\usepackage[T1]{fontenc}

\usepackage[utf8]{inputenc}

\usepackage{microtype}

\usepackage{inconsolata}

\usepackage{multirow}
\usepackage{subfig}
\makeatletter
\newcommand\notsotiny{\@setfontsize\notsotiny\@vipt\@viipt}
\makeatother

\usepackage{xcolor, soul}
\definecolor{ashgrey}{rgb}{0.80, 0.80, 0.80}
\sethlcolor{ashgrey}

\usepackage{graphicx}

\title{Adaptation of Biomedical and Clinical Pretrained Models to French Long Documents: A Comparative Study}

\author{Adrien Bazoge{\normalfont\textsuperscript{1,2}} \hspace{0.5cm} Emmanuel Morin{\normalfont\textsuperscript{2}}\\ {\bf Béatrice Daille}\textsuperscript{2} \hspace{0.5cm} \hspace{0.5cm} {\bf Pierre-Antoine Gourraud}\textsuperscript{1} \\ 
\textsuperscript{1}Nantes Université, CHU Nantes, Clinique des données, INSERM, CIC 1413, F-44000 Nantes, France \\ \textsuperscript{2}Nantes Université, École Centrale Nantes, CNRS, LS2N, UMR 6004, F-44000 Nantes, France   \\
\texttt{\{firstname.lastname\}@univ-nantes.fr}
}

\begin{document}
\maketitle
\begin{abstract}
 Recently, pretrained language models based on BERT have been introduced for the French biomedical domain. Although these models have achieved state-of-the-art results on biomedical and clinical NLP tasks, they are constrained by a limited input sequence length of 512 tokens, which poses challenges when applied to clinical notes. In this paper, we present a comparative study of three adaptation strategies for long-sequence models, leveraging the Longformer architecture.  We conducted evaluations of these models on 16 downstream tasks spanning both biomedical and clinical domains. Our findings reveal that further pre-training an English clinical model with French biomedical texts can outperform both converting a French biomedical BERT to the Longformer architecture and pre-training a French biomedical Longformer from scratch. The results underscore that long-sequence French biomedical models improve performance across most downstream tasks regardless of sequence length, but BERT based models remain the most efficient for named entity recognition tasks.
\end{abstract}

\section{Introduction}

Recent pretrained language models based on the Transformer architecture~\cite{vaswani2017attention} have succeeded in improving the field of natural language processing, enabling state-of-the-art performance across a broad range of tasks and forming the backbone of contemporary deep learning architectures.
Transformer-based models have been adapted to various domains, including biomedical and clinical. Initially, these adaptations used pretrained language models (PLMs) based on BERT~\cite{devlin-etal-2019-bert} like BioBERT~\cite{10.1093/bioinformatics/btz682}, ClinicalBERT~\cite{huang2019clinicalbert}, and PubMedBERT~\cite{10.1145/3458754}. Recently, there has been a notable shift towards the use of Large Language Models (LLMs) with models like Med-PaLM 2~\cite{Singhal2023Towards}, PMC-LLaMA~\cite{Wu2023PMC}, and MedAlpaca~\cite{Han2023MedAlpaca}.
Although LLMs have drawn the attention in recent years, several studies suggest that PLMs remain more suitable than LLMs for traditional NLP tasks, such as named entity recognition and text classification~\cite{he2023survey,ramachandran-etal-2023-prompt,labrak2023zeroshot,zhou-etal-2023-improving-transferability,yuan-etal-2023-zero}.

Due to the full self-attention mechanism, BERT-based models suffers from the limited input sequence length of 512 tokens when applied to clinical notes~\cite{gao2021limitations}.
Indeed, clinical documents are generally large documents that can contain up to several thousand words.
To address the constraints of modeling long texts with PLMs, the input sequence can be truncated to the initial 512 tokens or processed at the document-level using a sliding window of 512 tokens, with optional overlap.
However, such methods fail to consider long-range dependencies beyond 512 tokens, as they lack a self-attention mechanism that operates across segments processed in parallel.
Recently, various works have adapted the transformers' attention mechanism for long sequences by reducing its complexity~\cite{tay2021long}.
Some of these models were then adapted to the clinical domain for English~\cite{li2023comparative}. 

In this paper, we explore the adaptation of long-sequences pretrained models for French biomedical and clinical domains. 
The contributions of this article are the following:
\begin{itemize}
    \item The introduction of French pretrained language models based on the Longformer architecture for processing long texts in the biomedical and clinical domains. All models are freely available under the open-source Apache 2.0 license\footnote{\href{https://huggingface.co/abazoge/DrLongformer}{https://huggingface.co/abazoge/DrLongformer}}.
    \item A comparative study to analyze different pre-training strategies for the Longformer architecture. Our results demonstrate that further pre-training an English clinical model with French biomedical texts can yield better performance than converting a BERT model to the Longformer architecture or pre-training a French biomedical Longformer from scratch.
    \item We demonstrate that our long-sequence French biomedical models improve performance on a majority of biomedical and clinical tasks, regardless of sequence length, but BERT models remain the most efficient for NER tasks.
\end{itemize}

\section{Related works}

\subsection{Attention mechanism for long-sequence transformers}

The self-attention mechanism in Transformers is a key component of recent pretrained language models. It allows each token in an input sequence to interact independently and in parallel with all other tokens. However, the computational and memory demands of the Transformer's self-attention grow quadratically with the sequence length. This makes processing long sequences computationally expensive. Consequently, models like BERT and its variants are limited to input sequence lengths of 512 tokens, which can be limiting when working with long documents for specific tasks.

Various adaptation of the Transformers' self-attention mechanism have been proposed to reduce this complexity~\cite{tay2021long,10.1145/3530811}.
One of the possible approaches is sparse attention and involves using non-full attention patterns to reduce the scope of attention.
The sparse attention mechanisms implemented in various pretrained models often comprise several atomic patterns, including global attention, band attention (such as sliding window or local attention), dilated attention, random attention, and block local attention.

The Sinkhorn attention~\cite{tay2020sparse} combines a content-based block sparse attention along with block local attention.
Longformer~\cite{Beltagy2020Longformer} rely on a sliding window and a global attention.
In addition to sliding window and a global attention, BigBird~\cite{zaheer2020big} use an additional random attention to approximate full attention.
More recently, the LSG (Local Sparse Global) attention was introduced~\cite{10.1007/978-3-031-33374-3_35}. This attention relies on three components: block local attention, sparse attention to capture extended context and global attention.
All these attention mechanisms allow reducing the attention complexity from $O(n^2)$ to $O(nlogn)$ or $O(n)$.
Many other methods can improve the efficiency by replacing the self-attention matrix by a low-rank approximation, such as Linformer~\cite{wang2020linformer}, Performer~\cite{choromanski2020masked,choromanski2020rethinking}, Random Feature Attention~\cite{peng2021random}, and LUNA~\cite{ma2021luna}.

\subsection{Biomedical and clinical pretrained models}

BERT-based models~\cite{devlin-etal-2019-bert} have been extensively adapted for the biomedical and clinical domains. In English, various pretrained models have been introduced for the biomedical domain, such as BioBERT~\cite{10.1093/bioinformatics/btz682}, PubMedBERT~\cite{10.1145/3458754}, and SciBERT~\cite{beltagy2019scibert}. These models are primarily trained on scientific articles from sources like PubMed or Semantic Scholar. In the clinical domain, models like BlueBERT~\cite{peng-etal-2019-transfer} and ClinicalBERT~\cite{huang2019clinicalbert} stand out, both of which are trained on clinical narratives from the MIMIC database~\cite{johnson2016mimic}.

For languages other than English, the adaptation has been less extensive but nonetheless noteworthy. Models have been proposed for German~\cite{medbertde}, Portuguese~\cite{schneider-etal-2020-biobertpt}, Spanish~\cite{carrino2021biomedical}, and Turkish~\cite{turkmen2022bioberturk}. 
For French, multiple models have been introduced recently: AliBERT~\cite{berhe-etal-2023-alibert}, CamemBERT-Bio~\cite{touchent2023camembertbio}, and DrBERT~\cite{labrak-etal-2023-drbert}. They leverage biomedical data from various sources to pre-train their models, such as scientific articles, drug leaflets, and clinical cases.

For long documents, only the Longformer and BigBird models have been adapted with continual pre-training using the MIMIC database~\cite{johnson2016mimic}, resulting in the Clinical-Longformer~\cite{li2023comparative} and Clinical-BigBird~\cite{li2023comparative} models, respectively.
Beside English, no model has been adapted for other languages.

\section{Models pre-training}

In this section, we describe the pre-training modalities of our models.
We first introduce the pre-training strategies for the models' adaptation in Section~\ref{sec:pretraining_strategies}.
In Section~\ref{sec:pretraining_datasets}, we describe the dataset used for the models' pre-training.
Finally, in Section~\ref{sec:optimization}, we provide the technical details of the models' pre-training.

\subsection{Pre-training strategies}\label{sec:pretraining_strategies}
Prior studies on pre-training French biomedical language models have drawn different conclusions about the impact of pre-training strategies on downstream task performance.
\citet{labrak-etal-2023-drbert} showed with DrBERT that pre-training a BERT model from scratch is more effective than domain adaptation through continual pre-training from CamemBERT's weights, a French general model.
On the other hand, \citet{touchent2023camembertbio} demonstrate on NER tasks that CamemBERT-bio, a biomedical model also pretrained from the weights of CamemBERT, outperform DrBERT when evaluated with metrics other than macro F1.
These different conclusions suggest a reassessment of the findings.

In this study, we evaluate various pre-training strategies to adapt the Longformer architecture to the French biomedical domain and introduce three models:
\begin{itemize}
\item \texttt{DrBERT-4096}: Converting DrBERT into the Longformer architecture by replacing the weights of Longformer with those of DrBERT. 
\item \texttt{DrLongformer-FS}: Pre-training a model from scratch, including the subword tokenizer built from the pre-training dataset.
\item \texttt{DrLongformer-CP}: Continuing the pre-training of the clinical English model {Clinical-Longformer} on French biomedical data, while keeping the initial tokenizer.
\end{itemize}
Only English models can be used as a starting point for the continual pre-training strategy since no Longformer model is available for French.
All pre-training scripts for each strategy are available to reproduce the experiments\footnote{\href{https://github.com/abazoge/DrLongformer}{https://github.com/abazoge/DrLongformer}}.

\subsection{Datasets}\label{sec:pretraining_datasets}

For the pre-training of the models, we used NACHOS~\cite{labrak-etal-2023-drbert}, a corpus containing a wide range of biomedical data crawled from the web.
In total, the pre-training dataset consists of approximately 1.276 billion words, corresponding to 2,367,419 documents.
Each document was segmented into subword units using the tokenizer specific to each model, and then split into sequences of 4,096 subword units.
Sequences at the end of documents that were shorter than 512 tokens were not retained.
Those longer than 512 tokens were padded up to 4,096 tokens.

For the model pretrained from scratch, named DrLongformer-FS, we use a WordPiece~\cite{https://doi.org/10.48550/arxiv.1609.08144} tokenizer built from the pre-training dataset, with a vocabulary size of 50,265 tokens.
For the DrBERT-4096 and DrLongformer-CP models, the tokenizer is the same as DrBERT and Clinical-Longformer models respectively.

\subsection{Optimization \& Pre-training}\label{sec:optimization}

All pretrained models are based on the Longformer architecture (12 layers, 768 hidden dimensions, 12 attention heads, 149M parameters).
The models are trained on the Masked Language Modeling (MLM) task using HuggingFace library~\cite{https://doi.org/10.48550/arxiv.1910.03771} with a masking probability of 15\%.

For the DrLongformer-FS and DrLongformer-CP models, we performed around 50k steps with a batch size of 256, each sequence being filled with 4,096 tokens, making it possible to process around one million tokens per step.
The learning rate is gradually increased linearly over 10k steps, from zero to the initial learning rate of 5$\times$10$\textsuperscript{-5}$.
Models are trained on 128 Nvidia V100 32 GB GPUs for 20 hours.
We also use mixed-precision training (FP16)~\cite{https://doi.org/10.48550/arxiv.1710.03740} to reduce the memory footprint.

After adapting the DrBERT model to the Longformer architecture, resulting in the new model named DrBERT-4096, we pretrained it for an additional epoch on NACHOS, since DrBERT had already undergone pretraining on NACHOS.
We performed 13,300 steps over 22 hours using a single Tesla A100 GPU with a batch size of 8 and gradient accumulation over 4 steps, resulting in an effective batch size of 32.

\begin{table*}[ht]
\setlength\tabcolsep{5.0pt}
\scriptsize
\centering
\center
\resizebox{\textwidth}{!}{%
\begin{tabular}{cccrrrrr}
\hline
\textbf{Dataset} & \textbf{Task} & \textbf{Metric} & \textbf{Train} & \textbf{Validation} & \textbf{Test} & \textbf{\begin{tabular}[c]{@{}l@{}}Avg. seq.\\ length\end{tabular}} & \textbf{\begin{tabular}[c]{@{}l@{}}Max seq.\\ length\end{tabular}} \\ \hline

\multicolumn{2}{c}{\textit{Public datasets}} \\
\rule{0pt}{3ex}

ESSAI & POS Tagging & seqeval F1 & 5,072 & 725 & 1,450 & 22.55 & 135 \\
CAS Corpus & POS Tagging & seqeval F1 & 2,653 & 379 & 758 & 23.07 & 136 \\
CLISTER & Semantic textual similarity & EDRM/Spearman & 499 & 101 & 400 & 25.21 & 103 \\
DEFT-2020 & Semantic textual similarity & EDRM/Spearman & 498 & 102 & 410 & 45.14 & 118 \\
DEFT-2020 & Multi-class Classification & Weighted F1 & 460 & 112 & 530 & 90.19 & 254 \\
DEFT-2021 & Multi-label Classification & Weighted F1 & 118 & 49 & 108 & 404.78 & 1,710 \\
DEFT-2021 & NER & seqeval F1 & 121 & 46 & 108 & 404.78 & 1,710 \\
QUAERO - EMEA & NER & seqeval F1 & 13 & 12 & 15 & 1,006.42 & 1,549 \\
E3C - Clinical & NER & seqeval F1 & 146 & 22 & 81 & 356.39 & 714 \\
E3C - Temporal & NER & seqeval F1 & 56 & 8 & 17 & 354.86 & 685 \\
FrenchMedMCQA & MCQA & EMR/Hamming Score & 2,171 & 312 & 622 & 44.80 & 174 \\ 
FrenchMedMCQA & Multi-class Classification & Weighted F1 & 2,171 & 312 & 622 & 44.80 & 174 \\ 
MorFITT & Multi-label Classification & Weighted F1 & 1,514 & 1,022 & 1,088 & 226.33 & 1~425 \\
DiaMed & Multi-class Classification & Weighted F1 & 509 & 76 & 154 & 329.92 & 1,379  \\ \hline

\multicolumn{2}{c}{\textit{Private datasets}} \\
\rule{0pt}{3ex}

aHF classification & Binary Classification & Weighted F1 & 1,179 & 132 & 328 & 782.77 & 4,142 \\
aHF NER & NER & seqeval F1 & 358 & 51 & 103 & 929.78 & 1,992  \\ \hline

\end{tabular}%
}
\caption{Description, metrics and statistics of downstream biomedical and clinical NLP tasks. The average and max sequence length are words length.\label{fig:synthese_task_longformer}}
\end{table*}

\section{Downstream tasks evaluation}

We evaluate the long-sequence models on 16 downstream tasks.
First, we describe the datasets and their associated tasks in Section~\ref{sec:downstream_tasks}.
The performances of our models are compared to state-of-the-art French short-sequence biomedical models and English long-sequence models, which are presented in Section~\ref{sec:baselines_models}.
Finally, in Section~\ref{sec:fine_tuning_downstream}, we detail the fine-tuning procedure of the models on the downstream tasks along with the evaluation metrics.
All fine-tuning scripts are available to reproduce the experiments\footnote{\href{https://github.com/DrBenchmark/DrBenchmark}{https://github.com/DrBenchmark/DrBenchmark}}.

\subsection{Evaluation tasks}\label{sec:downstream_tasks}

The evaluation primarily focuses on datasets with long sequences, i.e., sequences longer than 250 words.
The sequence length in tokens depends on the tokenizer used. 
With an average tokenization of a word into two tokens, a sequence longer than 250 words would exceed the capacity of a short-sequence model like BERT. Therefore, long-sequence models are necessary to process such sequences with long-term dependencies.
Based on this selection criteria, the following biomedical datasets from DrBenchmark~\cite{labrak2024drbenchmark} were included in the evaluation: DEFT-2021, DiaMed, E3C, QUAERO-EMEA, MorFITT and the classification task from the DEFT-2020 dataset.
It is important to note that DrBenchmark's collection does not cover datasets derived directly from clinical practice. To address this gap and enhance the diversity of our data sources, we supplemented our evaluation with two tasks from an in-house clinical dataset focused on acute heart failure.
This initial set of tasks primarily consisted of document classification and NER.
To add more diversity in the evaluation tasks, we add POS tagging (CAS corpus and ESSAI), semantic textual similarity (CLISTER and DEFT-2020) and multiple-choice question answering (FrenchMedMCQA) tasks.

The descriptions and statistics of all datasets are presented in Table~\ref{fig:synthese_task_longformer}, and each dataset is further described below.

\paragraph{ESSAI}~\cite{dalloux_claveau_grabar_oliveira_moro_gumiel_carvalho_2021} is a corpus of 7,247 sentences taken from French clinical trials and clinical cases, annotated in 41 POS tags. The dataset was randomly split into 3 subsets of 70\%, 10\% and 20\% of the total data for training, validation and test respectively.

\paragraph{CAS corpus}~\cite{grabar:hal-01937096} contains sentences extracted from clinical cases that have been annotated for POS tagging with 31 classes. The dataset was randomly split into 3 subsets of 70\%, 10\% and 20\% of the total data for training, validation and test respectively.

\paragraph{CLISTER}~\cite{hiebel-etal-2022-clister-corpus} is a French clinical semantic textual similarity (STS) dataset of 1,000 sentence pairs manually annotated by several annotators, who assigned similarity scores ranging from 0 to 5 to each pair. The scores were then averaged together to obtain a floating-point number representing the overall similarity.

\paragraph{DEFT-2020}~\cite{cardon-etal-2020-presentation} contains clinical cases introduced in the 2020 edition of French Text Mining Challenge, called DEFT. The dataset is annotated for two tasks: (i) semantic textual similarity and (ii) multi-class classification. The first task aims at identifying the degree of similarity within pairs of sentences, from 0 (the less similar) to 5 (the most similar). The second task consists in identifying, for a given sentence, the most similar sentence among three sentences provided.

\paragraph{DEFT-2021}~\cite{grouin-etal-2021-classification} is a subset of 275 clinical cases taken from the 2019 edition of DEFT. This dataset is manually annotated in two tasks: (i) multi-label classification and (ii) NER. The multi-label classification task is annotated with 23 axes derived from Chapter C of the Medical Subject Headings (MeSH). The second task focuses on fine-grained information extraction for 13 entities, such as diseases, signs, or symptoms mentioned in the clinical cases.

\paragraph{QUAERO corpus}~\cite{Nvol2014TheQF} contains annotated entities and concepts for NER tasks. The dataset covers two text genres: EMEA, containing drug leaflets, and MEDLINE, containing scientific article titles, but only the EMEA subset was used in this study.
The 10 annotated entities correspond to the UMLS Semantic Groups~\cite{Lindberg-MIM1993}.
Due to the presence of nested entities, we simplified the evaluation process by retaining only annotations at the higher granularity level from the BigBio~\cite{fries2022bigbio} implementation, following the approach described in~\citet{touchent2023camembertbio}.
Owing to this choice translates into an average loss of 6.06\% of the annotations on the EMEA subset.

\paragraph{E3C}~\cite{Magnini2020TheEP} is a multilingual dataset of clinical narratives annotated for the NER task. It consists of two types of annotations: (i) clinical entities (e.g., pathologies), (ii) temporal information and factuality (e.g., events). While the dataset covers 5 languages, only the French portion is used in this study.
The dataset consists of two layers: layer 1 is manually annotated while layer 2 is semi-automatically annotated.
We randomly split the dataset with 70\% for training, 10\% for validation, and 20\% for testing, as explained in Table~\ref{table:sources-e3c}.

\begin{table}[!htb]
\setlength\tabcolsep{3.5pt}
\scriptsize
\centering
\begin{tabular}{|c|c|c|c|}
\hline
\textbf{Subset}   & \textbf{Train}      & \textbf{Validation}        & \textbf{Test} \\ \hline
\textbf{Clinical} & 87.38 \% of layer 2 & 12.62 \% of layer 2 & 100 \% of layer 1 \\ \hline
\textbf{Temporal} & 70 \% of layer 1    & 10 \% of layer 1    & 20 \% of layer 1 \\ \hline
\end{tabular}%
\caption{Description of the sources for the E3C subsets.}
\label{table:sources-e3c}
\end{table}

\paragraph{FrenchMedMCQA}~\cite{labrak:hal-03824241} is a Multiple-Choice Question Answering (MCQA) dataset for biomedical domain. It contains 3,105 questions coming from real exams of the French medical specialization diploma in pharmacy, integrating single and multiple answers. The first task consists of automatically identifying the set of correct answers among the 5 proposed for a given question. The second task consists of identifying the number of answers (between 1 and 5) supposedly correct for a given question.

\paragraph{MorFITT}~\cite{labrak:hal-04125879} is a multi-label dataset that has been annotated with specialties from the medical field. The dataset contains 3,624 abstracts from PMC Open Access. It has been annotated across 12 medical specialties, for a total of 5,116 annotations.

\paragraph{DiaMed} is a dataset of 739 clinical cases that have been manually annotated by several annotators, one of which is a medical expert, with 22 chapters of the International Classification of Diseases, 10th Revision (ICD-10). These chapters provide a general description of the type of injury or disease.

\paragraph{Acute heart failure (aHF) classification} This task consists of the classification of hospital stays reports according to the presence or absence of a diagnostic of acute heart failure. This corpus consists of 1,639 hospital stays reports from Nantes university hospital, which are labeled as positive or negative to acute heart failure.

\paragraph{Acute heart failure (NER)} This corpus contains 350 hospital stay reports from Nantes University Hospital. The reports are annotated with 46 entity types related to the following clinical information: cause of chronic heart failure, triggering factor for acute heart failure, diabetes, smoking status, heart rate, blood pressure, weight, height, medical treatment, hypertension and left ventricular ejection fraction. Overall, the corpus contains 6,116 clinical entities.

\subsection{Baselines models}\label{sec:baselines_models}

We describe the short and long-sequence pretrained models used as baselines in this study.

\paragraph{DrBERT}~\cite{labrak-etal-2023-drbert} is a French biomedical language model built using a from-scratch pre-training of the RoBERTa architecture on NACHOS, a French public biomedical corpus integrating 1.08 billion words (7.4 GB). 

\paragraph{CamemBERT-bio}~\cite{touchent2023camembertbio} is a French biomedical language model pretrained from the weights of CamemBERT~\cite{martin-etal-2020-camembert} on the French public corpus {biomed-fr} made of 413 million words (2.7 GB) and containing a wide range of data taken from the web.


\paragraph{Clinical-Longformer}~\cite{li2023comparative} is a clinical language model pretrained from the weight of the Longformer model on approximately 2 million clinical notes extracted from the MIMIC-III database~\cite{johnson2016mimic}. This model support input sequences up to 4,096 tokens long, which correspond to 8 times the input sequence limit of BERT models.

\subsection{Fine-tuning and evaluation metrics}\label{sec:fine_tuning_downstream}

All the models are fine-tuned regarding a strict protocol using the same hyperparameters for each downstream task. 
On every task, the batch size is 2 for Longformer based models, and 16 or 32 for BERT based models.
To ensure robustness and reliability in the evaluation, we experiment multiple learning rate: \{1$\times$10$\textsuperscript{-4}$, 1$\times$10$\textsuperscript{-5}$, 2$\times$10$\textsuperscript{-5}$, 5$\times$10$\textsuperscript{-5}$\}.
For each task, four runs are conducted for each learning rate, and we retain the results of the learning rate that achieves the best average over the four runs.
Finally, depending on the tasks and the models' architecture, the fine-tuning protocol and the evaluation metrics can vary.

\paragraph{Named entity recognition and POS tagging} For token classification tasks such as POS tagging and NER, a token classification head (a linear layer) is added on top of the hidden-states output.
BERT based models are fine-tuned and evaluated at sentence-level, whereas Longformer models are fine-tuned at document-level (or half document when it exceeds 4,096 tokens).
The models' performance is evaluated with the seqeval~\cite{seqeval} metric in strict mode with the IOB2 format. This metric provides an entity-level evaluation based on the first token of each entity, instead of a token-level basis. This approach offers a tokenizer-agnostic assessment, reducing any potential correlation between the models' performance and the tokenization process.

\paragraph{Text classification} For the document classification tasks, a sequence classification head is added on top of the pooled output.
Documents are truncated to the first 4,096 tokens to match the length limits of the Longformer models.
For the short-sequence models, we truncate the document to the first 512 tokens.
All classification tasks are evaluated with a weighted F1 score.

\paragraph{Semantic textual similarity} The CLISTER and DEFT-2020 datasets are fine-tuned as sequence classification with input formatted as \hl{\texttt{[CLS] <text1> [SEP] <text2> [EOS]}}.
No sequence truncation is needed, since the maximum sequence lengths of these datasets are less than 512 tokens.
Predictions are made by adding a linear layer on top of the pooled output.
The evaluation metrics used are the mean relative solution distance accuracy (EDRM), as defined by the original authors of the dataset~\cite{cardon-etal-2020-presentation}, and the Spearman correlation.

\paragraph{Multiple-choice question answering} The FrenchMedMCQA dataset is fine-tuned as sequence classification by adding a linear layer on top of the pooled output.
The input sequence is formatted as follows: \hl{\texttt{[CLS] <Question> [SEP] <Answer A> [SEP] <Answer B> [SEP] <Answer C> [SEP] <Answer D> [SEP] <Answer E> [EOS]}}.
As the maximum sequence lengths are less than 512 tokens, no truncation is needed.
For the output, the classification is multi-class with 31 classes corresponding to existing combinations of answers in the corpus. For example, if the correct answers are \texttt{A} and \texttt{B}, we consider the correct class being \texttt{AB}.
The models' performance is assessed using exact match ratio (EMR) and Hamming score.

\begin{table*}[ht!]
\centering
\resizebox{\textwidth}{!}{%
\begin{tabular}{|l|l|c|c|c|c|c|c|c|c|c|}
\hline

\textbf{Dataset} & \textbf{Task} & \textbf{DrBERT} & \textbf{CamemBERT-bio} & \textbf{Clinical-Longformer}  & \textbf{DrBERT-4096} & \textbf{DrLongformer-CP} & \textbf{DrLongformer-FS} \\ 
\hline 

 \multirow{1}{*}{ESSAI} & POS & {98.41*} & 97.39* & {98.54}* & \underline{98.60}* & \textbf{98.72} & 98.48* \\ 
\hline 

 \multirow{1}{*}{CAS} & POS & {96.93*} & 95.22* & 97.26* & \underline{97.55}* & \textbf{97.73} & 97.29* \\ 
\hline 

 \multirow{1}{*}{CLISTER} & STS & {0.62} / {0.57}* & 0.54* / 0.26* & {0.6} / \textbf{0.88} & \textbf{0.65} / 0.71 & {0.6} / \underline{0.86} & \underline{0.64} / 0.58* \\ 
\hline 

 \multirow{2}{*}{DEFT-2020} & STS & 0.72 / {0.81}* & 0.58* / 0.32* & 0.71 / \underline{0.85} & \underline{0.73} / 0.84 & \textbf{0.78} / \textbf{0.88} & 0.65 / 0.79* \\ 
  & CLS & 82.38 & \underline{94.78} & \textbf{95.59} & 87.48 & 94.27 & 36.57* \\ 
 
\hline 

 \multirow{2}{*}{DEFT-2021} & CLS & 34.15* & 17.82* & 27.73* & 36.98* & \textbf{51.23} & \underline{49.54} \\ 
 & NER & {60.44}* & \textbf{64.36} & 55.62* & 59.74* & \underline{63.73}* & 55.00* \\ 
\hline 

 \multirow{1}{*}{QUAERO-EMEA} & NER & {64.11} & \textbf{66.59} & 50.36* & 64.09 & 64.78 & \underline{66.31} \\ 
\hline 

 \multirow{2}{*}{E3C} & NER-Clinical & {54.45} & {56.96} & 50.88* & \textbf{57.20} & \underline{57.07} & 55.74 \\ 
 & NER-Temporal & {81.48}* & \textbf{83.44} & 79.11* & 81.98* & \underline{83.32} & 79.23* \\ 
\hline 

 \multirow{2}{*}{FrenchMedMCQA} & MCQA & \underline{31.07}* / {3.22}* & \textbf{35.3} / 1.45* & 28.55* / 1.45* & 28.2* / \textbf{3.54} & 28.62* / 1.45* & 23.5* / \underline{3.38} \\ 
 & CLS & 65.38 & {65.79} & 64.57 & \textbf{66.59} & \underline{66.0} & 64.42 \\ 
\hline 

 \multirow{1}{*}{MorFITT} & CLS & 68.70* & 67.53* & 62.98* & \underline{70.75}* & \textbf{71.95} & {69.94}* \\ 
\hline 

 \multirow{1}{*}{DiaMed} & CLS & \underline{60.45} & 39.57* & 39.91* & \textbf{61.17} & {54.82}* & 54.05* \\ 
\hline

 aHF classification & CLS & 83.25* & {83.47}* & 79.14* & \underline{85.22}* & \textbf{87.42} & 82.34* \\ \hline

aHF NER & NER & \underline{37.32} & 29.08* & 30.86* & \textbf{38.53} & {36.44}* & 34.81* \\ \hline

\hline
\end{tabular}%
}
\caption{Performance of the pretrained models on French biomedical and clinical downstream tasks. Best model in bold and second is underlined. Statistical significance is computed for each task using Student’s t-test with the best model as reference: * stands for p < 0.05.}
\label{table:results}
\end{table*}

\section{Results and Discussions}

The performance of the models on downstream tasks is presented in Table~\ref{table:results}.
We propose to analyze the results from two perspectives: (i) pre-training strategies and (ii) model architecture, considering both the maximum input sequence length and the attention mechanism.

\subsection{Impact of the pre-training strategies}

\paragraph*{DrBERT vs CamemBERT-bio} 

When comparing models solely based on the BERT architecture, we find that DrBERT and CamemBERT-bio have comparable performances. Specifically, each model outperforms in 8 of the 16 tasks evaluated. As described in Table~\ref{tab:average_perf}, CamemBERT-bio shows a slight advantage in NER tasks, with an average F1 score difference of 0.53. Conversely, DrBERT exhibits superior performance in classification tasks, yielding an average F1 score difference of 4.23. These observations indicate that there are ultimately no real differences on downstream tasks between training from scratch and continual pre-training for BERT models.

From an environmental impact perspective, continual pre-training appears to be the most responsible approach. In fact, the estimated carbon emission for pre-training CamemBERT-bio is 0.84 kg CO2eq. This estimation doesn't take into account the carbon emissions of the CamemBERT model itself. In contrast, the DrBERT model has emissions amounting to 36.9 kg CO2eq, which is approximately 44 times higher. Given that specialized domain models are built on existing architectures, it is more advantageous to continue pre-training from these models.

\paragraph*{Pre-training strategies for long-sequence models}

Contrary to what we observed with BERT biomedical models, using an existing model for pre-training—whether through conversion of BERT or continual pre-training of Longformer—yields better results than training a model from scratch.
Indeed, the DrLongformer-FS model never outperforms either DrBERT-4096 or DrLongformer-CP.
The results also highlights that DrLongformer-FS is better than Clinical-Longformer on only 10 of the 16 tasks.
However, it's important to note that none of the Longformer models mentioned in the literature have been pretrained from scratch.
For example, Clinical-Longformer was pretrained from the weights of Longformer, which in turn was pretrained from the weights of RoBERTa.
In the end, the DrLongformer-CP model underwent four pre-training phases, resulting in enhanced robustness due to the variety of data seen during the multiple pre-training phases.
For a fair comparison between DrLongformer-FS and other models, DrLongformer-FS would need a pre-training duration equivalent to the multi-phase pre-training experienced by the other models. However, this approach would be prohibitively costly, with no guarantee of significant performance improvement.

When comparing DrBERT-4096 and DrLongformer-CP, the results are more nuanced.
DrLongformer-CP, which was pretrained from the weights of Clinical-Longformer, achieves better results in 11 of the 16 tasks.
The DrLongformer-CP model appears to benefit from the clinical data from the MIMIC database used to pre-train the Clinical-Longformer model. This finding supports the conclusions drawn by~\citet{labrak-etal-2023-drbert} regarding the effectiveness of specialized knowledge transfer between languages, particularly when further pretraining an English clinical model on French biomedical data.
However, it remains uncertain whether the continual pretraining from a specialized English model offers more advantages than from a general French model, given the absence of a French Longformer model for such an experiment.
While the DrBERT-4096 model doesn't perform as well as the DrLongformer-CP, it's important to note that it offers a more cost-effective option for adapting a new architecture to a specialized domain. Despite its lower training cost, DrBERT-4096 still delivers performance levels that are comparably close to DrLongformer-CP across many tasks.

\subsection{Impact of the model architecture on downstream tasks} 

Regarding the architectures of pretrained models, the results show that Longformer models perform better than BERT models on 11 of the 16 tasks.
By analyzing the results according to the type of evaluation task from Table~\ref{table:results}, we can draw different conclusions.

\begin{table*}[ht!]
\centering
\resizebox{\textwidth}{!}{%
\begin{tabular}{|l|c|c|c|c|c|c|}
\hline
\textbf{Task} & \textbf{DrBERT}      & \textbf{CamemBERT-bio}  & \textbf{Clinical-Longformer} & \textbf{DrBERT-4096} & \textbf{DrLongformer-CP} & \textbf{DrLongformer-FS} \\ \hline
POS  & 97.67       & 96.31       & 97.90               & 98.08 &  \textbf{98.23}           & 97,89           \\ \hline
CLS  & {65.72}       & 61.49       & 61.65     & \underline{68.03}           & \textbf{70.95}           & 59.48           \\ \hline
NER  & 59.56       & {60.09}       & 53.37        & \underline{60.31}       & \textbf{61.07}           & 58.22           \\ \hline
STS  & \underline{0.67} / 0.69 & 0.56 / 0.29    & 0.66 / \textbf{0.87}   &  \textbf{0.69} / \underline{0.78}     & \textbf{0.69} / \textbf{0.87}     & 0.65 / 0.69     \\ \hline
MCQA & \underline{31.07} / {3.22} & \textbf{35.3} / 1.45 & 28.55 / 1.45 & 28.2 / \textbf{3.54} & 28.62 / 1.45 & 23.5 / \underline{3.38} \\
\hline 
\end{tabular}%
}
\caption{Average performance per task. Best model in bold and second is underlined.}
\label{tab:average_perf}
\end{table*}

\begin{table*}[h!]
\centering
\begin{tabular}{|l|rr|rr|rr|rr|}
\hline
 & \multicolumn{2}{c|}{\textbf{DEFT-2021}}& \multicolumn{2}{c|}{\textbf{DiaMed}} & \multicolumn{2}{c|}{\textbf{MorFITT}} & \multicolumn{2}{c|}{\textbf{aHF classification}} \\ \hline 
Sequence length & \multicolumn{1}{c|}{$\leq$ 388}   & \multicolumn{1}{c|}{$>$ 388} & \multicolumn{1}{c|}{$\leq$ 395}   & \multicolumn{1}{c|}{$>$ 395} & \multicolumn{1}{c|}{$\leq$ 406} & \multicolumn{1}{c|}{$>$ 406} & \multicolumn{1}{c|}{$\leq$ 400} & \multicolumn{1}{c|}{$>$ 400} \\ \hline
Documents in test set & \multicolumn{1}{c|}{78} & \multicolumn{1}{c|}{30} & \multicolumn{1}{c|}{112} & \multicolumn{1}{c|}{42} & \multicolumn{1}{c|}{1056} & \multicolumn{1}{c|}{32} & \multicolumn{1}{c|}{54} & \multicolumn{1}{c|}{274} \\ \hline
DrBERT & \multicolumn{1}{r|}{16.19} & 27.52 & \multicolumn{1}{r|}{52.22} & 47.37 & \multicolumn{1}{r|}{8.19} & \textbf{9.32} & \multicolumn{1}{r|}{5.64} & 14.70 \\ \hline
DrBERT-4096 & \multicolumn{1}{r|}{17.31} & 28.10 & \multicolumn{1}{r|}{\textbf{48.56}} & \textbf{36.27} & \multicolumn{1}{r|}{8.94} & 9.57 & \multicolumn{1}{r|}{9.68} & 13.87 \\ \hline
DrLongformer-CP & \multicolumn{1}{r|}{\textbf{14.31}} & \textbf{22.09} & \multicolumn{1}{r|}{58.53} & 38.73 & \multicolumn{1}{r|}{\textbf{7.94}} & 10.16 & \multicolumn{1}{r|}{\textbf{4.03}} & \textbf{7.55} \\ \hline
\end{tabular}
\caption{Error rates ($\downarrow$) of DrBERT, DrBERT-4096 and DrLongformer-CP on the test set of DEFT-2021 and aHF classification datasets. The selected word sequence length threshold for each dataset corresponds to the maximum length limit for DrBERT. Best model in bold.}
\label{tab:error_rate}
\end{table*}

\begin{figure*}[t]
\centering
\subfloat[aHF classification - Binary classification.]{\label{fig:attention_gavroche} \includegraphics[width=1\textwidth]{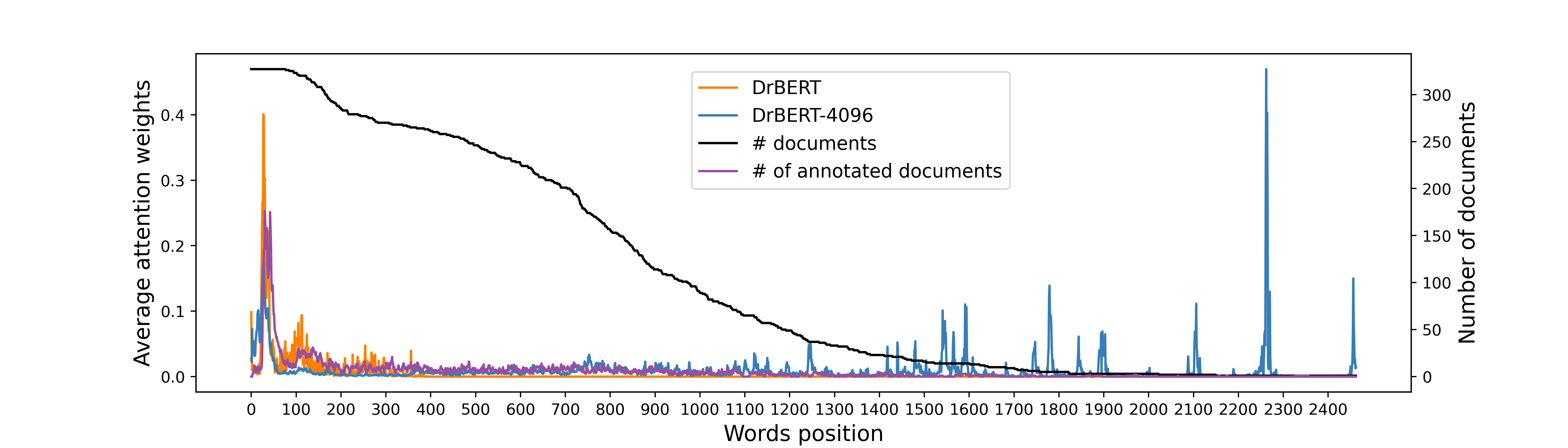}}%
\vspace{0.01cm}
\subfloat[DEFT-2021 - Multi-label classification.]{\label{fig:attention_deft2021} \includegraphics[width=1\textwidth]{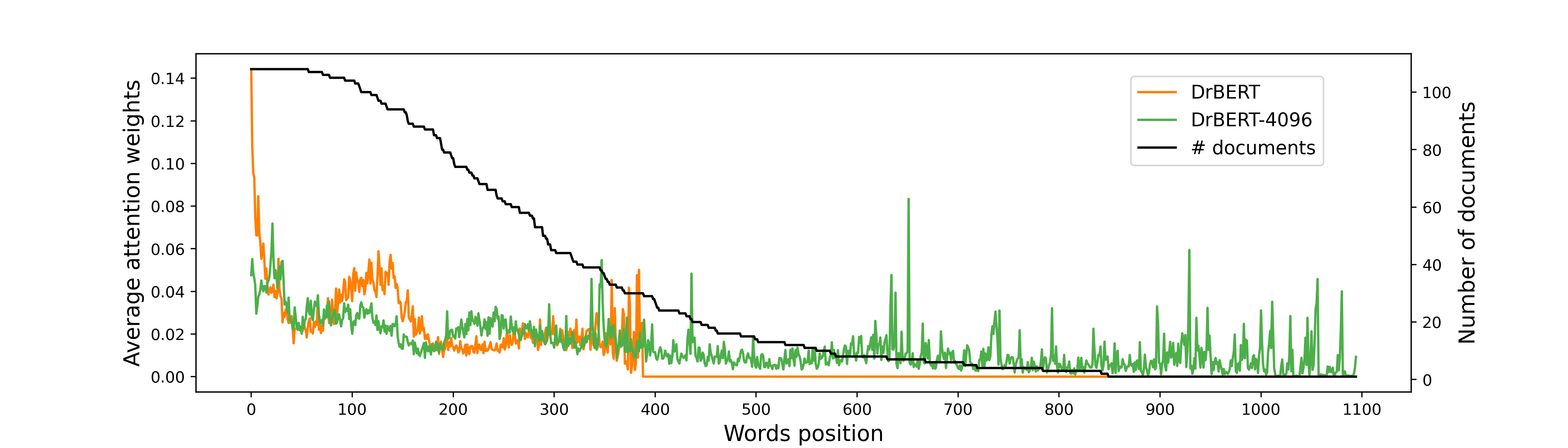}}%
\caption{Average attention weights for each word position on the test set of (a) the aHF classification dataset, and (b) the DEFT-2021 dataset. The attention weights are obtained from the [CLS] token used for classification by summing the attention weights of all attention heads in the last layer of the model. These weights represent the most important words after they have already incorporated contextual information from other words based off the 12 self-attention layers of the model. In (a) The aHF classification task is annotated at both document-level (Yes/No) and sequence-level. At the sequence-level, specific sequences of words that justify the document-level classification are annotated. This enables a comparison between human annotations and attention mechanisms of the models.}
\end{figure*}

\paragraph{Named entity recognition} We notice that the BERT models are more suited for NER tasks.
Indeed, even though DrBERT and CamemBERT-bio process NER tasks at the sentence level rather than the document level, they outperform the Longformer models in 3 of the 5 NER tasks.
This shows that global context is not necessarily required for NER tasks, and that the local context is generally sufficient for the evaluated datasets.
Although BERT models exhibit superior performance on individual NER tasks, Table~\ref{tab:average_perf} shows that DrLongformer-CP achieves a higher average F1 score across all tasks, demonstrating the model's robustness.

\paragraph{Document classification}
Sequence classification tasks appear to benefit from the Longformer's maximum sequence length of 4,096 tokens.
For tasks that involve sequences longer than 512 tokens, such as those in DEFT-2021, MorFITT, DiaMed, or aHF classification, the results show a clear advantage for Longformer models. On average, there is a gain of 6.25 points in the F1 score when comparing the best BERT model to the best Longformer model.
For these four datasets, we assessed the error rates for both short and long sequences. This analysis aimed to determine on which sequence type the Longformer models demonstrated improved performance.
In Table~\ref{tab:error_rate}, the error rates indicate that Longformer models outperform DrBERT on both short and long sequences. For short sequences, the average difference in error rates between DrBERT and the best-performing Longformer model is 1.85\%. For long sequences, this difference increases to 5.86\%.
Moreover, in tasks with shorter sequences, such as FrenchMedMCQA and DEFT-2020, the Longformer models still show a slight edge, with an average F1 score improvement of 0.81 over the BERT variants. This demonstrates a consistent advantage of Longformer models across a range of sequence lengths and classification tasks.

By analyzing the distribution of attention on the clinical reports from the aHF classification dataset in Figure~\ref{fig:attention_gavroche}, we observe that the attention is focused at the beginning of the document, in the same place as human annotations. Indeed, French clinical notes always start with the reason for the hospital stay, which is an important marker for the classification. This concentration of attention at the beginning of the documents also explains why BERT achieves good performance on this task, despite truncation to the first 512 tokens.
The attention peaks after the 1,500 word position correspond to the conclusions found at the end of clinical notes, which summarize the important elements of the clinical case. 
It is also important to note that the spikes are more prominent at the tail of the figure because the average is calculated over a smaller number of documents.
On the DEFT-2021 dataset, the model attention is evenly distributed across the entire document, as illustrated in Figure~\ref{fig:attention_deft2021}.
These two examples highlight the necessity of tailoring models to handle documents of different lengths and ensure that essential information is not missed.
\paragraph{Multiple-choice question answering}
On the multiple-choice question answering task from the FrenchMedMCQA corpus, BERT models appear to be more efficient with EMR scores exceeding 30\% compared to 28.62\% for the best Longformer models.
In Table 3, the two metrics often show opposite results. 
DrBERT-4096 and DrLongformer-FS exhibit EMR scores exceeding 3.0 but lower Hamming scores (below 30\%). In such instances, these models perform well in answering questions with a single correct answer but struggle when confronted with questions that have multiple answer variations to consider. 
On the contrary, while CamemBERT-bio excels at providing partial answers to questions, achieving a Hamming score of 35.3\%, it encounters more challenges when tasked with questions that have a single, definitive answer.

\paragraph{Semantic textual similarity} The reported results between the models are very close on DEFT-2020 and CLISTER datasets, with neither BERT nor Longformer distinctly standing out. In terms of Spearman correlation, there is no significant difference between BERT and Longformer.

\section{Conclusion}

In this work, we proposed a comparative study of adaptation strategies for French biomedical and clinical long-sequence models.
Specifically, we have shown that there is no significant difference in performances between pre-training from scratch and continual pre-training for models based on BERT architecture. 
On the other hand, for larger models, converting or further pre-train an existing model yields better performance than training from scratch.
We have also demonstrated for the continual pre-training strategy that building a specialized French model from a specialized English model is possible.
Using an English specialized model becomes a good alternative when there is no French-specific model available for the desired architecture.
These findings suggest that domain transfer can be achieved cross-lingually through multi-phase adaptive pre-training.
Furthermore, the lack of open French clinical data can be compensated by combining English clinical data from the MIMIC database with French biomedical data.

\section*{Limitations}


The comparative study conducted here is difficult to reproduce, since it required a considerable amount of computational resources.
Approximately 5,000 hours of GPU were necessary to pre-train the three new models introduced in this study.
For the evaluation of these models across the evaluation tasks, 1,536 fine-tuning sessions were conducted, for a total of 4,400 GPU hours.
The fine-tuning cost is higher for Longformer models than for BERT models. As the model size increases, there's a greater need to reduce the batch size.
In total, approximately 9,400 hours of GPU computation were used for this work.
The environmental cost is equivalent to 2,434.6 KWh or 138.77 kg CO2eq.

\section*{Acknowledgements}
This work was performed using HPC resources from GENCI-IDRIS (Grant 2023-AD011013715R1) and CCIPL (Centre de Calcul Intensif des Pays de la Loire). This work was financially supported by ANR AIBy4 (ANR-20-THIA-0011).

\bibliography{anthology,custom}
\bibliographystyle{acl_natbib}




\end{document}